\documentclass[11pt]{aime}

\title{Interface-Conditioned Message Passing for Polymer Property Prediction}

\author[1]{Varun Ajith}
\author[1,2,*]{Rocío Mercado}

\affiliation[1]{AI Laboratory for Molecular Engineering (AIME), Department of Computer Science and Engineering, Chalmers University of Technology \& University of Gothenburg, Gothenburg, Sweden}
\affiliation[2]{Science for Life Laboratory (SciLifeLab), Gothenburg, Sweden}

\contribution[*]{Corresponding author}  

\correspondence{\email{rocio.mercado@chalmers.se}}
\codeavailability{\url{https://github.com/varunajith/IC-wD-MPNN}}
\date{\today}

\abstract{Accurate polymer property prediction requires representations that capture both the chemical environments of constituent monomers and the connectivity of polymer chains. Although graph neural networks have been adapted to model stochastic polymer connectivity, incorporating higher-level chemical context while retaining the underlying graph topology remains an important challenge. Here, we introduce an interface-conditioned extension of the weighted directed message passing neural network that enriches atom representations at polymerization interfaces through message regeneration, incorporating contextual information from recurring chemical motifs. We refer to our approach as IC-wD-MPNN and evaluate it on electron affinity (EA) and ionization potential (IP) prediction using the VIPEA dataset, and on glass-transition temperature (T\textsubscript{g}) prediction using the PolyMetriX dataset. On VIPEA, the proposed formulation reduces mean RMSE by 36.3\% for EA and 40.7\% for IP relative to the baseline wD-MPNN on a comparable parameter budget. In contrast, no overall improvement is observed for T\textsubscript{g} prediction.
Ablation studies show that regenerating interface messages using contextualized atom embeddings already provides substantial gains, while incorporating motif information into these embeddings yields a further improvement. A random-message regeneration control further indicates that these gains depend on selecting messages originating from polymerization interfaces, rather than simply regenerating an equivalent number of messages. These results highlight the potential of localized interface conditioning to enrich polymer representations and improve the prediction of electronic properties.
}

\begin{document}
\maketitle

\section{Introduction}
\label{sec:intro}

Polymers are used across applications ranging from structural composites and membranes to battery electrolytes and organic semiconductors \cite{tran2024design}. Their versatility stems from the many ways in which monomer chemistry, composition, and chain architecture can be combined to produce materials with diverse properties \cite{aldeghi2022graph}. Machine learning (ML) offers a way to explore this large design space by predicting material properties and screening candidate polymers at a much lower computational cost than evaluating each candidate via \textit{ab initio} calculations or experimentally \cite{duan2022machine}. These predictions can help identify promising materials, uncover relationships between polymer structure and properties, and guide the selection of candidates for further investigation \cite{ge2025machine}. Beyond property prediction, ML can also support inverse design, where the goal is to generate new polymer structures that meet desired property targets \cite{yue2025benchmarking}. The usefulness of these approaches, however, depends on how well the underlying representations capture the chemical and structural features that govern polymer properties, especially in the low-data regimes generally typical of polymer informatics.

Polymer-specific graph neural network (GNN) frameworks have emerged in recent years, for instance extending message passing to account for the stochastic connectivity between repeat units, allowing models to represent variations in polymer composition and architecture \cite{aldeghi2022graph}. In parallel, molecular representation learning has explored ways to incorporate information at multiple levels of chemical structure, including recurring substructures, or motifs \cite{fan2026hierarchical, hong2026hierarchical,ma2026hyg} and recent work has also extended these ideas to polymer-specific hierarchical representations \cite{sun2026hipoly,erdogan2026s}. Here, we introduce an interface-conditioned extension of the weighted directed message passing neural network (wD-MPNN \cite{aldeghi2022graph}), termed IC-wD-MPNN. The method learns representations of recurring chemical motifs from contextualized atom embeddings and uses them to condition the representations of polymerization-interface atoms. These atoms are directly involved in connecting adjacent repeat units, and messages originating from them account for approximately 40\% of all directed messages on average across batches, with the proportion varying between 30\% and 50\% across individual polymer graphs. Regenerating these messages provides a way to introduce motif-derived information at the interfaces and propagate it through subsequent message passing.

We evaluate IC-wD-MPNN on electron affinity (EA) and ionization potential (IP) prediction using the VIPEA \cite{aldeghi2022graph} dataset, and on glass-transition temperature (\(T_g\)) prediction using the PolyMetriX \cite{kunchapu2025polymetrix} dataset. On VIPEA, the model reduces mean RMSE relative to the baseline wD-MPNN by 36.3\% for EA and 40.7\% for IP across the monomer-based evaluation folds, with EA improving in 8 and IP in 7 of the 9 monomer-based evaluation folds. These gains are not uniform across folds or individual predictions, and the evaluation on PolyMetriX does not show an overall improvement for \(T_g\). Ablation studies show that regenerating interface messages using contextualized atom embeddings already accounts for a substantial part of the improvement, while incorporating motif information into these embeddings provides a further, smaller reduction in RMSE. An additional random-message regeneration control shows that regenerating the same number of randomly selected messages does not reproduce the gains from interface atom-specific regeneration. Together, these findings support the hypothesis that polymerization interfaces provide particularly informative locations for message updates.


\section{Background}

\subsection{Graph representations for polymer property prediction}

Graph neural networks (GNNs) are widely used for molecular property prediction \cite{han2023himgnn,ma2026hyg,zang2023hierarchical}, representing molecules as graphs in which atoms are nodes and chemical bonds are edges. Message-passing neural networks learn atom representations by iteratively aggregating information from neighboring atoms, allowing each representation to capture its local chemical environment. For polymers, graph-based methods extend this framework to represent repeat units, monomer structures, and their connectivity when predicting material properties \cite{aldeghi2022graph,antoniuk2022representing,kimmig2026structure}. \citet{medina2026graph} reviewed recent applications of GNNs to polymer characterization and property prediction, covering the range of graph representations and learning approaches used in this area.

\citet{aldeghi2022graph} introduced a weighted directed message-passing approach (wD-MPNN) for polymer property prediction, using stochastic graph representations to account for polymer connectivity. Subsequent work has explored different ways of representing polymer structure, addressing aspects that are not necessarily captured by a single graph formulation. \citet{antoniuk2022representing} represented polymers as periodic graphs and learned descriptors from these representations, providing an alternative way to encode repeating polymer connectivity. \citet{queen2023polymer} developed PolymerGNN, a multitask graph-learning framework that predicts multiple polymer properties from monomer composition. Other approaches have focused on more compact string-based representations. \citet{ahn2024enhancing} proposed HAPPY, which abstracts repeat units into functional groups, ring structures, and connectors, with Topo-HAPPY extending this representation to capture additional topological information. \citet{kazemi2026descriptor} compared molecular-descriptor-based random forests with GNNs operating on 2D polymer graphs for seven copolymer properties, finding that the relative performance of these representations varies across properties. More recently, approaches have begun to explicitly incorporate the hierarchical organization of polymers. \citet{kimmig2026structure} developed a hierarchical graph model in which nodes represent monomer units, incorporating molecular-mass-distribution information to represent polymer samples as statistical ensembles. \citet{sun2026hipoly} further extended this direction with HiPoly, which integrates atom-, motif-, and monomer-level representations within a unified graph architecture, allowing information to be exchanged across these levels of polymer structure. Collectively, these approaches reflect a growing interest in representing polymers beyond their atomic structures and connectivity, incorporating information about larger chemical substructures and polymer organization.

Self-supervised learning offers another direction for improving polymer representations, particularly in settings where labeled property data are limited. Rather than relying exclusively on property-specific supervision, these approaches aim to learn transferable structural information from polymer data that can subsequently be used for downstream prediction tasks. \citet{gao2024self} explored this through pretraining a wD-MPNN using node-, edge-, and graph-level tasks on polymer graphs encoding monomer combinations, stoichiometry, and chain architecture, before fine-tuning the model for property prediction. \citet{piccoli2026joint} extended this direction by applying a joint embedding predictive architecture (JEPA) to stochastic polymer graphs. Instead of reconstructing masked input features, their approach learns to predict target-view embeddings from context-view embeddings. These studies illustrate different strategies for learning polymer representations from structural information beyond directly supervised property prediction. Beyond representation learning for prediction, learned models have also been used to generate new polymer candidates. \citet{vogel2025inverse} developed a graph-to-string variational autoencoder for inverse design, generating copolymer candidates while accounting for monomer stoichiometry and chain architecture. Together, these developments point towards a broader role for learned polymer representations, extending from property prediction and the discovery of structure-property relationships to the generation and design of new polymer candidates.

Despite these advances, representing chemical information at different levels of structural detail remains an important gap in polymer GNNs. While atom-level message passing can capture local chemical environments, recurring substructures are not represented explicitly as distinct chemical units. Introducing motif-level information therefore offers a way to provide the model with an additional representation of recurring chemical patterns, which may be particularly useful when labeled polymer data are limited. This has motivated the use of motif-based and hierarchical representations in molecular learning.

\subsection{Motif-based and hierarchical molecular representations}

Motif-based molecular representations capture chemical information at the level of recurring substructures, complementing the local environments encoded by individual atoms and bonds. \citet{jin2020hierarchical} explored this concept in molecular generation through a hierarchical graph encoder-decoder that uses structural motifs as building blocks for constructing molecular graphs. \citet{zhang2021motif} introduced Motif-based Graph Self-Supervised Learning (MGSSL), which uses BRICS-based molecular fragmentation and motif-generation tasks to incorporate substructure information into self-supervised pretraining for molecular property prediction.

Subsequent work has focused on explicitly integrating atom- and motif-level information into molecular representations. \citet{zang2023hierarchical} proposed Hierarchical Molecular Graph Self-supervised Learning (HiMol), which introduces motif and graph-level nodes into an augmented molecular graph and combines hierarchical graph encoding with multi-level self-supervised pretraining. Similarly, \citet{han2023himgnn} developed HimGNN, which uses separate atom- and motif-level graph encoders, with cross-attention and contextual rescaling to combine their representations for property prediction. More recently, \citet{wang2025motif} introduced Uni-Motif, a framework that separates motif identity from its structural context and incorporates both into different GNN architectures without requiring an explicit hierarchical graph. Collectively, these studies demonstrate different ways of incorporating chemically meaningful substructures into molecular learning, ranging from motif-based generation and pretraining to their explicit integration within graph architectures.

\subsection{Positioning of the present work}
\label{sec:background-positioning}

Applying motif-based representations to polymers raises an additional question concerning the role of motifs within a repeat unit. Some motifs contain atoms involved in polymerization interfaces, through which adjacent repeat units are connected, whereas others occur entirely within the repeat unit. We hypothesize that this positional distinction may influence the usefulness of motif information for polymer property prediction. In particular, incorporating motif information selectively at polymerization interfaces may provide a more effective way of capturing the relationship between local chemical structure and polymer connectivity than treating all motif occurrences uniformly.

To investigate this, we introduce an interface-conditioned formulation of the wD-MPNN, termed IC-wD-MPNN. Unlike approaches that explicitly construct hierarchical graph representations, such as HiPoly \cite{sun2026hipoly}, our approach incorporates motif information within the existing atom-level wD-MPNN framework. Motif representations are derived from contextualized atom embeddings and used to condition messages originating from polymerization-interface atoms, without introducing additional motif-level nodes or changing the underlying graph topology. We evaluate this formulation across different polymer properties and evaluation folds, with additional experiments examining the effect of these design choices. This allows us to assess both the contribution of interface-specific motif conditioning and whether the learned motif representations provide benefits beyond conventional atom- and bond-level features.

\section{Methods}
\label{sec:methods}

\begin{figure*}[h!]
    \centering
    \includegraphics[
        width=0.96\textwidth,
        height=2.85in,
        keepaspectratio
    ]{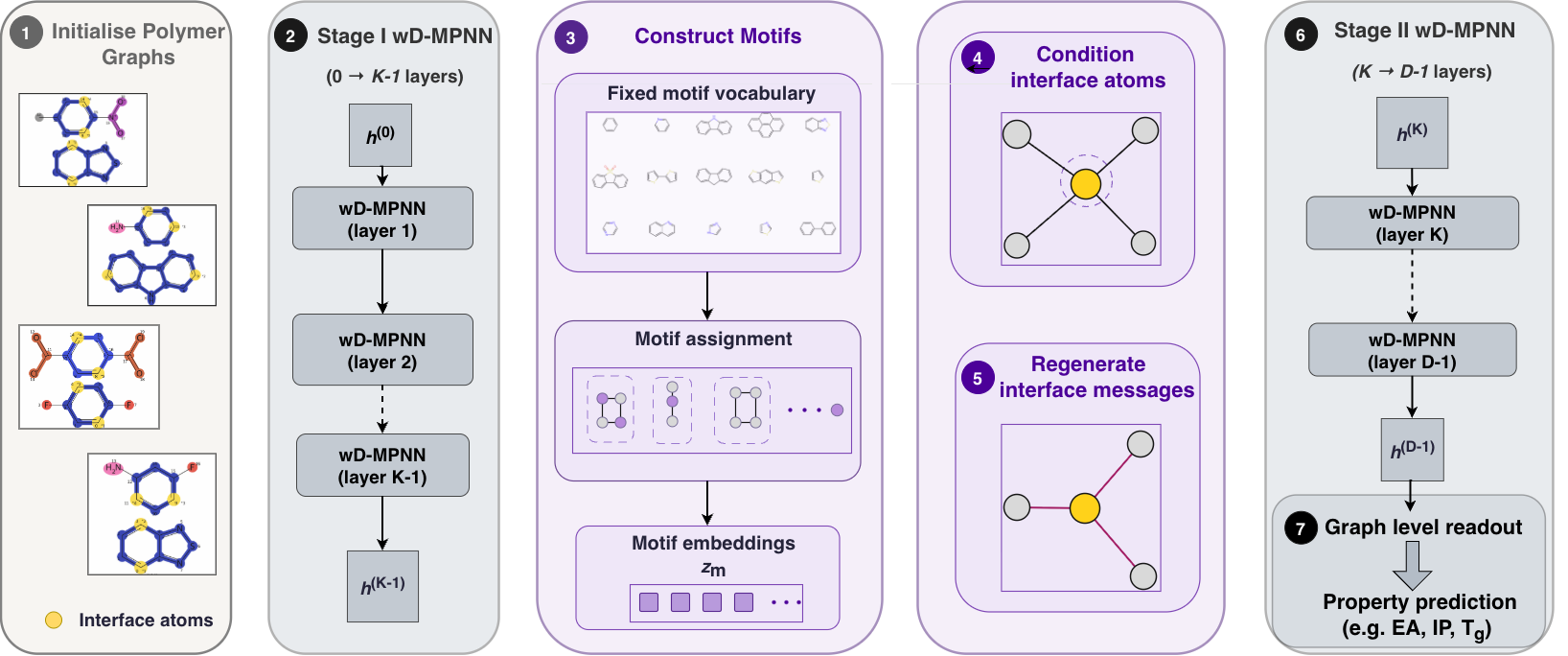}
    \caption
    {Overview of the IC-wD-MPNN formulation. The original polymer stochastic graph (1) is processed by an initial stage of message passing as implemented by the baseline wD-MPNN (2). The motif-based interface-conditioned module (blocks 3–5) constructs motif embeddings, uses them to enrich polymerization-interface atoms, and regenerates outgoing messages from these atoms. Message passing then continues on the stochastic graph (6), followed by graph-level readout and property prediction (7), as implemented by the  baseline wD-MPNN. The numbered blocks (1–7) indicate the sequential workflow. Detailed equations comparing the baseline and proposed formulation are provided in Appendix Table \ref{tab:model-comparison}.
    }
    
    \label{fig:pipeline}
\end{figure*}

\subsection{Data}
\label{sec:data}

We evaluate the proposed method on the VIPEA polymer dataset introduced by \citet{aldeghi2022graph}, using electron affinity (EA) and ionization potential (IP) as prediction targets. The dataset comprises 42,966 copolymers constructed by combining nine monomers in group A with 682 monomers in group B, across different chain architectures and monomer stoichiometries. EA and IP values were calculated using extended tight-binding (xTB) methods on polymer oligomers and subsequently corrected using reference DFT calculations. For direct comparison with the baseline wD-MPNN, we use the same nine monomer-A-based folds introduced by \citet{aldeghi2022graph}. Each fold holds out polymers containing monomer A. Note, however, that group A monomers also occur in group B; therefore, the split evaluates transfer across monomer combinations rather than generalization to wholly unseen monomer chemistry.

We use the PolyMetriX~\cite{kunchapu2025polymetrix} dataset to evaluate glass-transition temperature ($T_g$) prediction. It contains 7,367 unique homopolymer structures represented as canonicalized polymer SMILES (PSMILES), paired with curated experimental $T_g$ values collected from multiple literature sources. We evaluate the model using 16 homopolymer-class folds, providing a complementary assessment to the monomer-based evaluation on VIPEA.

\subsection{Polymer representation and model architecture}
\label{sec:polymer-architecture}

We build on the weighted directed message passing neural network (wD-MPNN) formulation \cite{aldeghi2022graph} for polymer property prediction. The wD-MPNN represents polymers as stochastic molecular graphs, accounting for variability in monomer composition, stoichiometry, and chain architecture. We extend this framework by deriving motif-level representations from intermediate atom embeddings and incorporating them into subsequent atom-level message passing at polymerization interfaces.

We denote the polymer graph as $G=(V,E)$, where $V$ and $E$ represent the sets of atoms and directed bonds, respectively. Following the wD-MPNN formulation, message passing is performed along directed bonds, with each message associated with a source atom and a neighboring destination atom. At each update, the message associated with a directed bond is computed by aggregating messages arriving at its source atom, excluding the message associated with the reverse direction of that bond. This prevents immediate backtracking along the same bond.

We denote the total message-passing depth by $D$ and define the insertion depth of the interface-conditioned module as
\begin{equation}
    K = \left\lfloor \frac{D}{2} \right\rfloor,
    \label{eq:motif-depth}
\end{equation}
placing the module midway through the encoder. At this depth, the atom representations have undergone sufficient message passing to capture local chemical environments, providing the contextualized embeddings used to construct motif representations (Fig.~\ref{fig:pipeline}). These motif embeddings are then incorporated into interface atom representations, and their outgoing messages are regenerated. The remaining $D-K$ layers propagate the enriched representations through the polymer graph, allowing the conditioned information to influence subsequent message passing.

After the final message-passing layer, we aggregate the atom representations using the original polymer readout and pass the resulting embedding to the property prediction head. The architecture therefore combines atom-level representations learned from the polymer graph with motif-level representations derived from them. The following sections describe the construction of the motif vocabulary, motif representation learning, and interface conditioning.

\subsection{Motif representation learning}
\label{sec:motif-representation}

The repeat unit is decomposed into chemically meaningful motifs using a vocabulary of 112 distinct Murcko scaffolds \cite{bemis1996properties}, supplemented by functional-group and backbone motifs. These motifs are derived from recurring structural patterns in the VIPEA dataset \cite{aldeghi2022graph}, and the resulting vocabulary is fixed before model training. Motif assignment follows a priority-based procedure in which scaffold matches are considered in descending order of heavy-atom count, giving larger scaffolds precedence over their constituent substructures. Atoms assigned to a scaffold are excluded from further matching, after which the remaining atoms are assigned to functional-group motifs and finally to backbone motifs. This produces a deterministic, non-overlapping assignment in which every atom belongs to exactly one motif. For the T\textsubscript{g} experiments, the same motif vocabulary constructed from VIPEA is reused without modification or adaptation to the PolyMetriX dataset. The most common Murcko scaffolds in the resulting vocabulary are illustrated in the Appendix (Fig.~\ref{fig:top-murcko-scaffolds}).

Let $\mathcal{M}=\{M_1,M_2,\ldots,M_{|\mathcal{M}|}\}$ denote the set of motif instances in the polymer graph, where each $M_m\subseteq V$ contains the atoms assigned to motif instance $m$. The mapping $\phi:V\rightarrow\mathcal{M}$ identifies the motif instance containing each atom. While motif assignments are determined by the structural decomposition, their representations are learned from the atom embeddings produced during the first $K$ message-passing layers.

For each motif $M_m$, attention-based pooling combines the contextualized atom embeddings obtained after the first $K$ message-passing layers:
\begin{equation}
    \mathbf{z}_m =
    \operatorname{AttnPool}
    \left(
        \left\{\mathbf{h}_i^{(K)}:i\in M_m\right\}
    \right).
    \label{eq:motif-pooling}
\end{equation}
Here, $\mathbf{h}_i^{(K)}$ denotes the embedding of atom $i$ after $K$ layers, and $\mathbf{z}_m$ is the resulting motif embedding. The learned attention weights allow different atoms to contribute unequally to the representation. Since pooling operates on contextualized atom embeddings, each motif representation reflects both its chemical structure and the local environment captured by the wD-MPNN. These embeddings are computed once at the conditioning depth and retained throughout the subsequent message-passing layers; they are not recomputed from the updated atom representations in later layers.

\subsection{Interface motif conditioning}
\label{sec:interface-conditioning}

Let $\mathcal{I}\subseteq V$ denote the set of polymerization-interface atoms identified during graph construction. For each interface atom $v_i\in\mathcal{I}$, we incorporate the embedding of its assigned motif into its contextual atom representation through direct addition:
\begin{equation}
    \widetilde{\mathbf{h}}_i^{(K)}
    =
    \mathbf{h}_i^{(K)}+\mathbf{z}_{\phi(i)},
    \qquad v_i\in\mathcal{I}.
    \label{eq:interface-enrichment}
\end{equation}
Here, $\mathbf{h}_i^{(K)}$ is the atom embedding after the first $K$ message-passing layers, and $\mathbf{z}_{\phi(i)}$ is the embedding of the motif containing that atom. This conditioning is applied only to polymerization-interface atoms, providing a localized way to introduce motif information at the points where repeat units connect within the polymer representation.

The enriched representations are then used to regenerate the outgoing directed messages from interface atoms. For a directed bond $(i,j)\in E$, let $\mathbf{m}_{ij}^{(K)}$ denote the message after the first $K$ layers, with $v_i$ as the source atom and $v_j$ as its neighbor. The regenerated message is defined as
\begin{equation}
    \widetilde{\mathbf{m}}_{ij}^{(K)}
    =
    \begin{cases}
        \tau\!\left(
        \mathbf{W}_r
        \left[
        \widetilde{\mathbf{h}}_i^{(K)}
        \,\Vert\, \mathbf{e}_{ij}
        \right]
        \right), & v_i\in\mathcal{I},\\[6pt]
        \mathbf{m}_{ij}^{(K)}, & v_i\notin\mathcal{I},
    \end{cases}
    \label{eq:interface-message}
\end{equation}
where $\mathbf{e}_{ij}$ denotes the directed bond features, $\Vert$ denotes concatenation, $\mathbf{W}_r$ is a learnable projection matrix, and $\tau$ is the activation function used in the message update. The proportion of interface-originating messages is calculated as the number of directed messages whose source atom belongs to the polymerization-interface set \(I\), divided by the total number of directed messages in the graph. Thus, messages originating from interface atoms are recomputed using their conditioned representations, whereas messages from other atoms retain their existing values.

The regenerated messages replace their corresponding pre-conditioning messages and initialize the remaining $D-K$ message-passing layers. These layers follow the original wD-MPNN update mechanism, including its transition-weighted message aggregation, with the addition of the skip connections in each update; these use the regenerated messages $\widetilde{\mathbf{m}}_{ij}^{(K)}$ in place of the initial messages $\mathbf{m}_{ij}^{0}$ (Appendix Table~\ref{tab:model-comparison}). The underlying stochastic connectivity and transition probabilities therefore remain unchanged, while subsequent message passing propagates the conditioned information beyond the interface atoms and into the surrounding polymer graph.

\subsection{Training details}
\label{sec:training}

The baseline and interface-conditioned wD-MPNN were trained using the same optimization settings across both datasets, with three random seed initializations for each evaluation fold. Both models were trained for 30 epochs, using a two-epoch learning-rate warm-up. The learning rate increased from $10^{-4}$ to $10^{-3}$ before returning to $10^{-4}$. To maintain a comparable parameter budget, the baseline hidden dimension was set to 400, compared with 300 for the proposed model, resulting in approximately 594k and 540k trainable parameters, respectively. The remaining model and training hyperparameters are summarized in Appendix Table~\ref{tab:hyperparameters}.

\subsection{Evaluation metrics}
\label{sec:metrics}

Predictive performance is evaluated using the root mean squared error (RMSE) between predicted and reference property values. RMSE penalizes larger prediction errors more strongly than smaller ones, making it useful when substantial deviations from reference property values are particularly undesirable in material property prediction. Lower RMSE values indicate better predictive accuracy. For VIPEA, performance is evaluated across nine monomer-based folds, while PolyMetriX uses 16 polymer-class folds. Each fold is evaluated using three random seed initializations. Fold-level RMSE values are used to assess performance across the held-out groups, while the variation across seeds captures sensitivity to random initialization.

%

\section{Results}
\label{sec:results}

We first evaluate the proposed model on electronic properties, namely electron affinity (EA) and ionization potential (IP), which are the target properties of the VIPEA dataset \cite{aldeghi2022graph}. We then examine its performance on glass-transition temperature (T\textsubscript{g}), a bulk thermodynamic property evaluated on the  PolyMetriX dataset using a polymer-class split evaluation \cite{kunchapu2025polymetrix}. This progression allows us to assess the model across different properties and evaluation settings.

\begin{figure}[h!]
    \centering
    \includegraphics[
        width=\linewidth
    ]{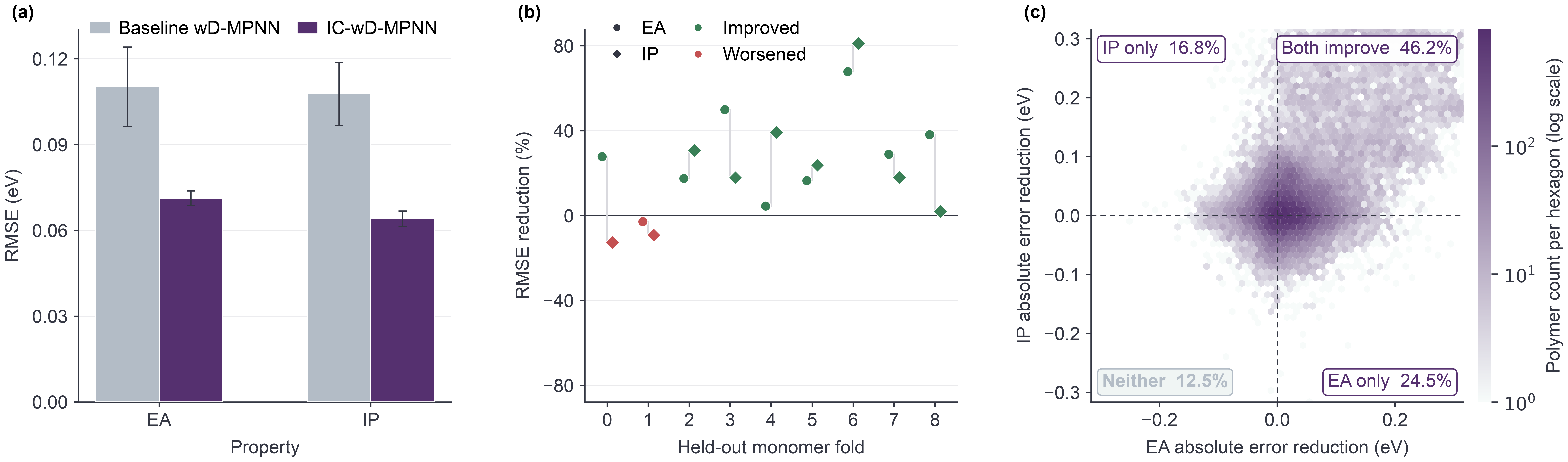}
    \label{fig:EA_IP_composite}
    \caption{Comparison of the baseline wD-MPNN and the interface-conditioned wD-MPNN (IC-wD-MPNN) for electron affinity (EA) and ionization potential (IP) prediction on the VIPEA dataset. (a) Overall test RMSE for both models, averaged across three random seeds; error bars indicate the standard deviation across seeds. Lower RMSE indicates better predictive performance. (b) Fold-wise percentage reduction in RMSE for IC-wD-MPNN relative to the baseline, averaged across seeds. Positive values (green) indicate improved performance, while negative values (red) indicate degraded performance. (c) Joint distribution of paired EA and IP prediction-error reductions across all 42,966 polymer configurations evaluated in the nine monomer-based folds. For each polymer configuration, absolute prediction errors are averaged across three random seeds, and the reduction is calculated as the baseline error minus the IC-wD-MPNN error. Positive values indicate lower prediction errors with IC-wD-MPNN. Hexagon color represents the number of datapoints per bin (log scale). Quadrant annotations indicate the percentage of datapoints showing improvement in both properties, EA only, IP only, or neither.}
    \label{fig:main_results}
\end{figure}

\subsection{Performance across electronic properties}
\label{sec:results-ea}

Figure~\ref{fig:main_results} compares the baseline wD-MPNN and IC-wD-MPNN for EA and IP prediction on the VIPEA dataset. As shown in Fig.~\ref{fig:main_results}(a), for EA prediction, the mean RMSE decreases from 0.110 to 0.070 eV, corresponding to a 36.3\% reduction. Similarly, for IP prediction, the RMSE decreases from 0.108 to 0.064 eV, representing a 40.7\% reduction. The fold-wise results in Fig.~\ref{fig:main_results}(b) show that the improvement is not uniform across the monomer-based splits: although most folds benefit, some show a modest increase in RMSE. In particular, fold 6 shows the largest reduction in RMSE for both EA and IP.

Figure~\ref{fig:main_results}(c) examines changes in prediction error at the individual-polymer level. Both EA and IP errors decrease for 46.2\% of polymers; a further 24.5\% show a reduction in EA error only, and 16.8\% show a reduction in IP error only. For the remaining 12.5\%, the baseline wD-MPNN has lower errors for both properties. Thus, the aggregate improvements are consistent with a broad tendency toward lower prediction errors across the test polymers, rather than being driven solely by a small subset of samples.

\subsection{Fold-wise performance across EA, IP and T\textsubscript{g}}
\label{sec:results-electronic}

\begin{figure}[h!]
    \centering

    \includegraphics[
        width=\linewidth
    ]{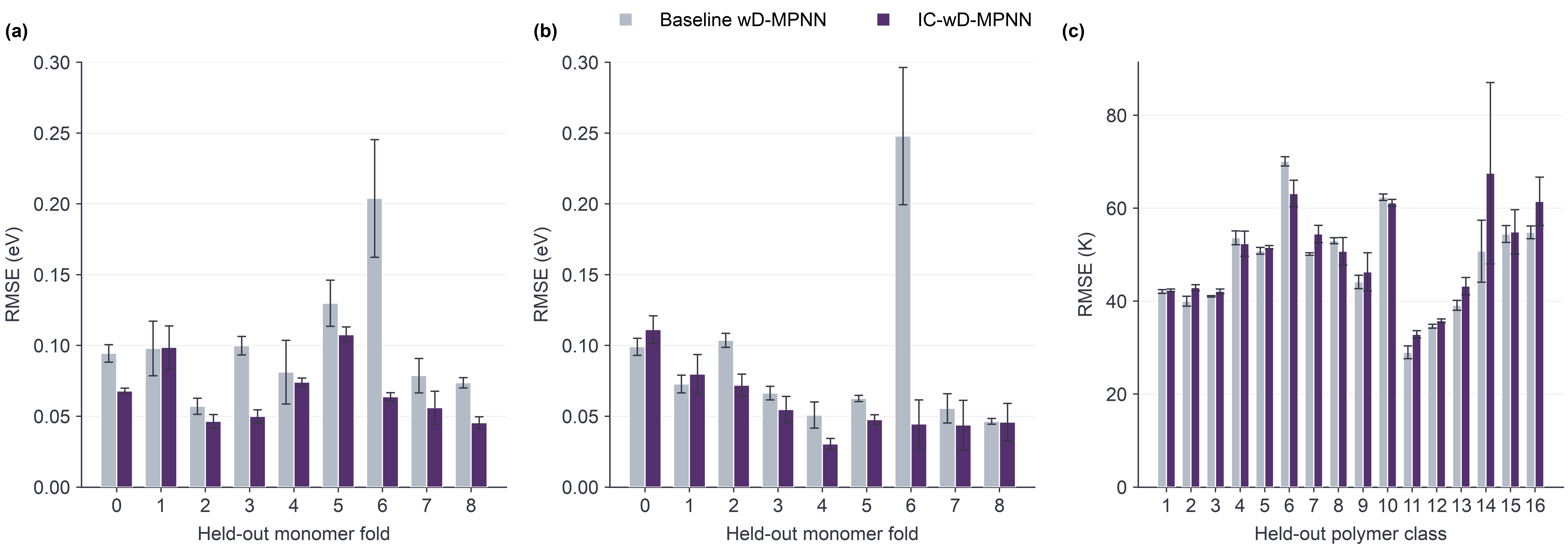}

    \caption{
    {Fold-wise RMSE comparison between the baseline wD-MPNN and the proposed formulation for (a) electron affinity (EA) and (b) ionization potential (IP) prediction on the VIPEA dataset, and (c) glass-transition temperature (T\textsubscript{g}) prediction on the PolyMetriX dataset. EA and IP results are reported across nine monomer-based folds, while T\textsubscript{g} results are reported across 16 polymer-class folds. Bars represent the mean test RMSE across three seeds, with error bars indicating $\pm 1$ standard deviation. Lower RMSE indicates better predictive performance.
    }
    }
    \label{fig:EA-IP-TG-results}
\end{figure}

Figure~\ref{fig:EA-IP-TG-results} provides a fold-wise comparison of RMSE for EA and IP prediction on VIPEA and T\textsubscript{g} prediction on PolyMetriX. For EA, the proposed formulation achieves lower RMSE in eight of the nine folds, with the largest reduction observed in Fold~6. The IP results show a similar  pattern: the proposed model performs better in most folds, although the improvements are less consistent in the first two folds. These results indicate that the aggregate gains observed for EA and IP are reflected across a majority of the individual monomer-based splits, while also revealing variation in the extent of improvement across folds. Notably, Fold 6 determines most of the aggregate gain; without it, the pooled error reduction drops from about 36\% to 21\% for EA, and 41\% to 9\% for IP. 

In contrast, the T\textsubscript{g} results show a more mixed pattern across the 16 polymer-class folds. The proposed formulation achieves lower RMSE in only a subset of folds, while the baseline performs better in several others, including some with relatively large differences between the two models. This is consistent with the aggregate results, where the baseline achieves slightly lower overall RMSE than the proposed formulation. The comparison therefore suggests that the improvements observed for EA and IP do not transfer uniformly to T\textsubscript{g} prediction, highlighting the dependence of model performance on the target property and evaluation setting.

\subsection{Ablation studies}
\label{sec:ablation}

Table~\ref{tab:ablation-results} presents the ablation results for EA and IP prediction on the VIPEA dataset. The two-stage variant without interface conditioning, in which neither motif enrichment nor interface-message regeneration is applied, achieves performance close to the baseline wD-MPNN. This suggests that the change in residual-message formulation alone has little effect on predictive performance under the present setup. Replacing the learned motif embeddings with fixed, non-trainable random vectors assigned to each motif type results in lower RMSE than the baseline, but does not match the performance of the variant without motif enrichment. A substantial reduction in RMSE is already observed when outgoing messages from interface atoms are regenerated using their contextualized atom embeddings, without motif enrichment. Incorporating learned motif embeddings into these representations provides a further reduction in RMSE, with the full IC-wD-MPNN achieving the lowest errors for both EA and IP. Overall, the results indicate that interface message regeneration accounts for a substantial part of the improvement, while learned motif representations provide an additional benefit.

To further examine whether the improvements depend on where messages are regenerated, we performed an additional control in which, for each polymer graph, the same number of directed messages as in the interface-conditioned setting was selected randomly from the full directed-message pool, rather than specifically from polymerization-interface atoms. The selected messages were regenerated using their contextualized atom embeddings, without motif enrichment. As shown in Table~\ref{tab:ablation-results}, it's performance remains close to the baseline for EA, while providing a slight improvement for IP. In contrast, regenerating messages specifically from interface atoms using contextualized embeddings alone achieves substantially lower errors for both properties. This comparison suggests that the improvements are not simply a consequence of regenerating additional messages, but depend on selecting messages associated with polymerization interfaces.

Additional ablation analyses examine the effects of message-passing depth, motif-vocabulary dropout, and motif pooling. The depth comparison evaluates IC-wD-MPNN against baseline wD-MPNNs with different message-passing depths, while the pooling comparison evaluates mean and attention pooling for constructing motif embeddings (Appendix, Tables~\ref{tab:depth-ablation} and~\ref{tab:motif-pooling}). The effect of motif-vocabulary dropout and the comparison of model performance across training-set fractions are also reported in the Appendix (Table~\ref{tab:motif-vocabulary-dropout} and Fig.~\ref{fig:training_data_fraction}).

\begin{table}[t]
    \centering
    \caption{Ablation results for electron affinity (EA) and ionization potential (IP) prediction on the VIPEA dataset. Values represent mean RMSE ± standard deviation across three random seeds, with predictions pooled across the nine evaluation folds for each seed. All values are reported in eV, with the best-performing variant for each property highlighted in bold.
    }
    \label{tab:ablation-results}
\begin{tabular}{lcc}
    \toprule
    \textbf{Model / Ablation} & \textbf{EA RMSE (eV)} & \textbf{IP RMSE (eV)} \\
    \midrule
    Baseline wD-MPNN
        & $0.110 \pm 0.013$
        & $0.108 \pm 0.011$ \\
    Two-stage wD-MPNN (no conditioning)
        & $0.109 \pm 0.012$
        & $0.106 \pm 0.014$ \\
    Interface message regeneration (random motif embeddings)
        & $0.097 \pm 0.019$
        & $0.099 \pm 0.021$ \\
    Random message regeneration (atom embeddings)
        & $0.110 \pm 0.004$
        & $0.093 \pm 0.005$ \\
    Interface message regeneration (atom embeddings)
        & $0.071 \pm 0.002$
        & $0.067 \pm 0.003$ \\
    \textbf{IC-wD-MPNN}
        & $\mathbf{0.070 \pm 0.002}$
        & $\mathbf{0.064 \pm 0.003}$ \\
    \bottomrule
\end{tabular}
\end{table}

\subsection{Runtime analysis}
\label{sec:runtime}

The full IC-wD-MPNN requires approximately three times the training time of the baseline wD-MPNN under the same experimental setup: around 3 hours per run, compared with approximately 1 hour for the baseline. This additional cost reflects the interface-conditioning operations introduced by the proposed formulation.

%


\section{Discussion}
\label{sec:discussion}

\subsection{Interface conditioning and electronic-property prediction}
\label{sec:discussion-electronic}

The improvements in EA and IP prediction support the hypothesis motivating IC-wD-MPNN: that messages originating from polymerization-interface atoms carry particularly relevant information for representing the connections between adjacent repeat units. Although they constitute only a small subset of the atoms in the polymer graph, they can account for a considerably larger fraction of the directed messages. This reflects the stochastic graph construction in wD-MPNN, where probabilistic connections between repeat units give rise to directed messages originating from interface atoms. Thus, the proportion of interface-originating messages is not directly determined by the fraction of atoms classified as interface atoms. By updating these messages, the model can refine how information associated with polymerization interfaces is propagated through the polymer graph.

The random-message control provides empirical support for this design choice. Regenerating the same number of randomly selected messages does not reproduce the improvements obtained through interface-specific regeneration. This suggests that the gains are not simply a consequence of additional message updates, but depend on targeting polymerization interfaces. The ablation results further show that interface regeneration using contextualized atom embeddings already accounts for much of the improvement, while incorporating learned motif embeddings provides an additional reduction in error.

The additional benefit of motif enrichment may reflect the value of capturing chemical context beyond individual atom environments. Motifs provide a representation of recurring chemical substructures that may be relevant to electronic properties such as EA and IP. However, the contribution of individual motifs and the specific chemical effects responsible for the improvements have not yet been established. Overall, the results support the original motivation for interface conditioning, while suggesting that motif-derived context can further improve the representations formed at these locations.

\subsection{Fold-wise performance and monomer correspondence}
\label{sec:discussion-transfer}

The improvement offered by interface conditioning varies across the nine monomer-based folds, suggesting that its benefits depend partly on the composition of the held-out monomer group. Fold 6 is particularly interesting in this regard. The baseline exhibits relatively high prediction errors for both EA and IP in this fold, while IC-wD-MPNN achieves a substantial reduction in both. Notably, Fold 6 is the only fold in which the held-out group-A monomer has no identical counterpart among the group-B monomers; such a counterpart is present in the other eight folds. This observation raises the possibility that the proposed model can produce enhanced performance even when the held-out group-A monomer identity is not represented in group B. In this sense, the improvement in Fold 6 may reflect the model’s ability to exploit shared structural features across different monomer combinations, rather than relying on the presence of an identical monomer in the training data. However, since this behaviour is observed in only one fold, it should be interpreted as suggestive rather than conclusive.

\subsection{Property dependence: electronic properties and glass-transition temperature}
\label{sec:discussion-property-transfer}

The results for T\textsubscript{g} differ from those for EA and IP. IC-wD-MPNN does not improve the overall RMSE on the PolyMetriX dataset~\cite{kunchapu2025polymetrix}. One possible explanation is that the three properties place different demands on the polymer representation. EA and IP are closely related to electronic structure and the chemical environments of constituent groups. In contrast, T\textsubscript{g} is a bulk material property influenced by several factors, including chain flexibility, segmental mobility and intermolecular interactions. Capturing local chemical substructures alone may therefore be insufficient to represent the broader structural effects that influence glass-transition behavior \cite{baker2022cooperative}. The result does not rule out the usefulness of motif information for T\textsubscript{g}, but suggests that localized motif conditioning, in its current form, does not capture all the information needed for this prediction task.

The T\textsubscript{g} experiments also use the motif vocabulary developed for VIPEA, without adapting it to the chemical structures in PolyMetriX. It remains to be investigated whether a vocabulary better suited to this dataset would change the results. More generally, the findings raise two questions for future work: how motif vocabularies should be defined for different polymer datasets, and how local chemical representations can be combined with information about larger-scale polymer structure. Addressing these questions could provide a route toward more expressive hierarchical polymer representations.

\subsection{Limitations and future work}
\label{sec:discussion-limitations}

The current approach has several limitations. The fixed motif vocabulary may not capture chemical structures beyond those represented in the dataset, while the non-overlapping assignment prevents atoms from contributing to multiple motifs that share them. In addition, motif conditioning is localised to polymerization-interface atoms, with each interface atom conditioned only on the motif to which it is assigned. Consequently, other motifs elsewhere in the polymer graph that do not contain the interface atom do not directly contribute to its conditioning. The additional interface-conditioning operations increase training time to approximately three times that of the baseline wD-MPNN. Although the proposed model operates on the same underlying graph topology, this additional computational cost should be considered when applying the approach to larger datasets or more extensive hyperparameter searches. Finally, we note that IC-wD-MPNN was not directly benchmarked against other motif-aware or hierarchical architectures, such as \citet{sun2026hipoly} and \citet{kimmig2026structure} , as these approaches address different tasks, use different datasets and representation frameworks. Future work could explore more flexible motif vocabularies, allow overlapping motif assignments, and investigate ways to incorporate motif information beyond the polymerization interfaces.

The evaluation also shows that the benefits of motif conditioning depend on the target property. While the model improves aggregate prediction accuracy for EA and IP on VIPEA, it does not improve overall RMSE for T\textsubscript{g} on PolyMetriX. Evaluating the approach across additional polymer datasets and properties, and assessing whether the motif vocabulary should be adapted to each target property, could help clarify the settings in which interface-conditioned representations are most effective.


\section{Conclusion}
\label{sec:conclusion}

In this work, we introduced an interface-conditioned extension of the weighted directed message passing neural network (IC-wD-MPNN) for polymer property prediction. The approach derives latent representations of recurring chemical motifs and uses them to enrich atom-level representations at polymerization interfaces before regenerating their outgoing messages. This provides a way to introduce explicit substructure information into polymer message passing without requiring a separate motif-level graph or changes to the underlying graph topology.

Evaluation on the VIPEA dataset showed lower aggregate RMSE for both electron affinity (EA) and ionization potential (IP), with gains varying across monomer-based folds and individual polymers. The ablation results showed that message regeneration using contextualized interface atom embeddings accounts for a substantial part of these improvements, while incorporating motif information provides an additional reduction in RMSE. Furthermore, a random-message regeneration control showed that regenerating the same number of randomly selected messages did not reproduce the gains from interface-specific regeneration, providing empirical support for the hypothesis underlying the architecture. On the separate PolyMetriX dataset, the approach did not improve overall glass-transition temperature (T\textsubscript{g}) prediction, indicating that its benefits are not uniform across properties. Together, these findings highlight the potential of interface-conditioned message passing as a means of incorporating motif-derived context into atom-level polymer representations, particularly for predicting electronic properties.


\section*{Author contributions}

VA: Conceptualization, methodology, software, data curation, formal analysis, investigation, validation, visualization, writing – original draft, and writing – review \& editing. RM: Conceptualization, methodology, supervision, funding acquisition, resources, and writing – review \& editing.


\begin{ack}
VA acknowledges funding from the European Research Council (ERC) under the European Union’s Horizon Europe research and innovation programme (POLYGEN project, Project ID: 101220355). RM acknowledges funding provided by the Wallenberg AI, Autonomous Systems, and Software Program (WASP), supported by the Knut and Alice Wallenberg Foundation. Computational resources were provided by the National Academic Infrastructure for Supercomputing in Sweden (NAISS), funded by the Swedish Research Council.

\textit{Funded by the European Union. Views and opinions expressed are, however, those of the author(s) only and do not necessarily reflect those of the European Union or the European Research Council Executive Agency. Neither the European Union nor the granting authority can be held responsible for them.}
\end{ack}

\section*{Notes}
The authors declare no competing financial interest.

\section*{Software and data availability}
Code is available at \url{https://github.com/varunajith/IC-wD-MPNN}. All data (VIPEA, PolyMetriX) can be downloaded from the original publications, cited in the Methods.


\bibliographystyle{unsrtnat}
\bibliography{references}


%
%

\newpage
\appendix
\section{Appendix}
\label{app:motifs}
\begin{figure}[h]
    \centering
    \begin{minipage}{0.62\linewidth}
        \centering
        \includegraphics[width=\linewidth]{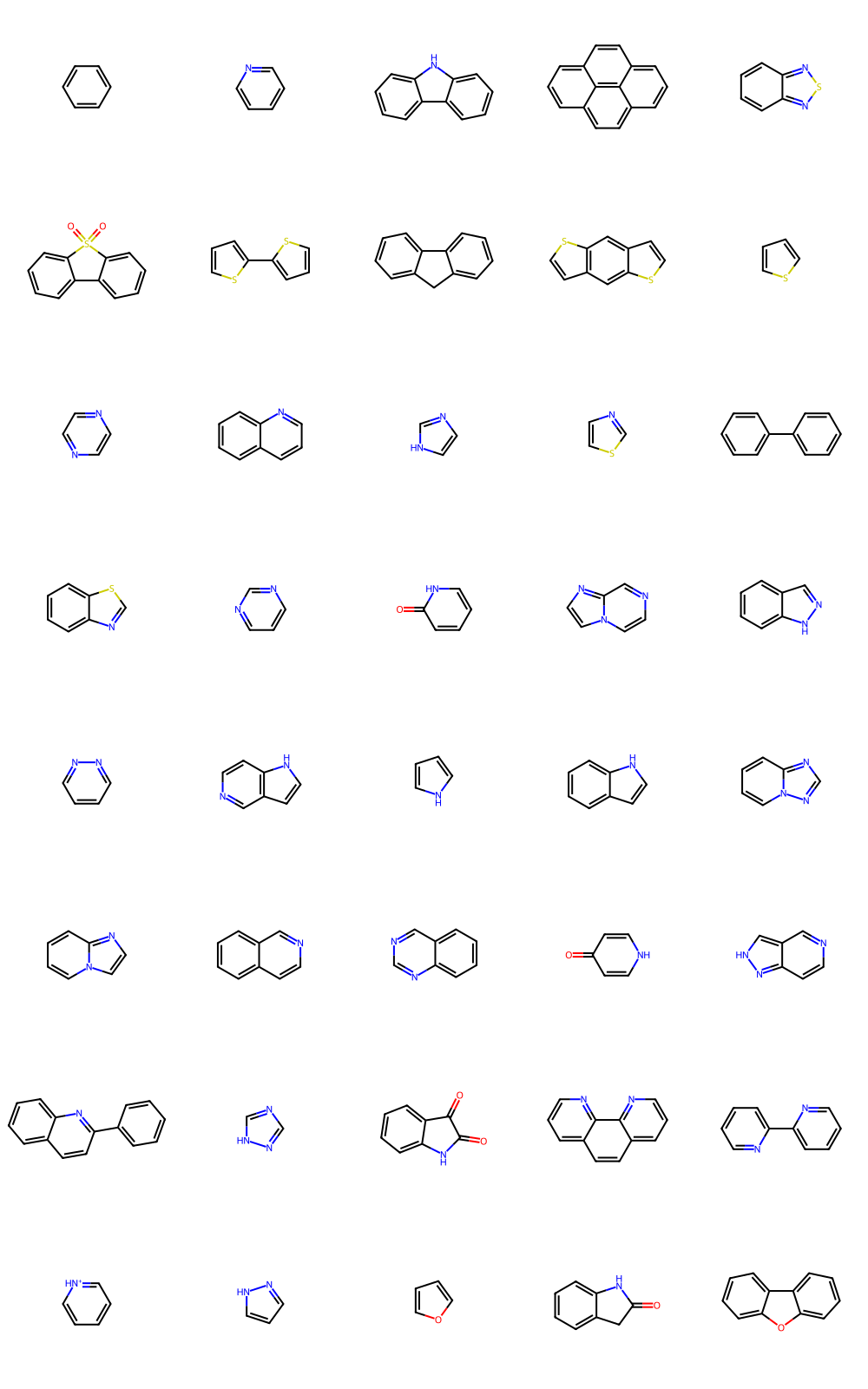}
        \caption{Examples of recurring Murcko scaffolds used in the predefined motif vocabulary.}
        \label{fig:top-murcko-scaffolds}
    \end{minipage}
\end{figure}

Figure~\ref{fig:top-murcko-scaffolds} shows the 40 most common of the 112 Murcko scaffolds that make up the fixed motif vocabulary. The vocabulary is dominated by the fused and linked (hetero)aromatic cores characteristic of conjugated photocatalyst chemistry (e.g., thiophene and benzodithiophene, carbazole and fluorene, benzothiadiazole, dibenzothiophene-\textit{S},\textit{S}-dioxide and pyrene) which makes a scaffold-level abstraction a natural fit for this dataset. Scaffolds are matched in descending order of heavy-atom count, so the larger fused systems shown here take priority over their own substructures during assignment.

\vspace{5mm}

\newpage

Table~\ref{tab:hyperparameters} lists the hyperparameter settings used for the baseline wD-MPNN and the proposed formulation. These settings provide the training and architectural details needed to reproduce the experiments.

\begin{table}[h]
\centering
\caption{
{Key model hyperparameters used for the baseline and interface-conditioned wD-MPNN. The two models use the same architecture and training settings except for the baseline hidden dimension, which was increased from 300 to 400 to obtain a comparable parameter budget.}
}
\label{tab:hyperparameters}
\begin{tabular}{lcc}
\toprule
\textbf{Hyperparameter} & \textbf{Baseline wD-MPNN} & \textbf{IC-wD-MPNN} \\
\midrule
Hidden size & 400 & 300 \\
Message-passing depth ($D$) & 3 & 3 \\
Activation & ReLU & ReLU \\
Dropout & 0.0 & 0.0 \\
FFN layers & 2 & 2 \\
Aggregation & Mean & Mean \\
Epochs & 30 & 30 \\
Warmup epochs & 2 & 2 \\
Initial learning rate & $10^{-4}$ & $10^{-4}$ \\
Maximum learning rate & $10^{-3}$ & $10^{-3}$ \\
Final learning rate & $10^{-4}$ & $10^{-4}$ \\
\midrule
Total parameter count & $\sim$594k & $\sim$540k \\
\bottomrule
\end{tabular}
\end{table}

Table~\ref{tab:depth-ablation} compares EA and IP prediction performance for the baseline wD-MPNN at message-passing depths of \(D=4,5,\) and \(6\) (hidden dimension = 300) with the proposed formulation. The results demonstrate that the gains from motif-enriched interface conditioning are not reproduced by the tested deeper baselines.

\begin{table}[t]
    \centering
    \caption{Effect of message-passing depth on EA and IP prediction performance on the VIPEA dataset. Values represent mean RMSE ± standard deviation across three random seeds, with predictions pooled across the nine evaluation folds for each seed. All values are reported in eV, with the best-performing depth for each property highlighted in bold.}
    \label{tab:depth-ablation}
    \begin{tabular}{lcc}
        \toprule
        \textbf{Model} & \textbf{EA RMSE (eV)} & \textbf{IP RMSE (eV)} \\
        \midrule
        wD-MPNN ($D=4$)
            & $0.093 \pm 0.012$
            & $0.085 \pm 0.013$ \\
        wD-MPNN ($D=5$)
            & $0.093 \pm 0.011$
            & $0.083 \pm 0.011$ \\
        wD-MPNN ($D=6$)
            & $0.105 \pm 0.011$
            & $0.072 \pm 0.011$ \\
        \midrule
         \textbf{IC-wD-MPNN}
            & $\mathbf{0.070 \pm 0.002}$
            & $\mathbf{0.064 \pm 0.003}$ \\
        \bottomrule
    \end{tabular}
\end{table}

Table~\ref{tab:motif-vocabulary-dropout} examines the effect of reducing the motif vocabulary on EA and IP prediction. Increasing the dropout fraction is associated with higher RMSE for both properties. Relative to the full vocabulary, 50\% dropout produces a modest increase in error, while 75\% dropout results in a larger degradation. This trend suggests that retaining a broader set of motifs is beneficial for the predictive performance of the proposed formulation.

\begin{table}[t]
    \centering
    \caption{Motif-vocabulary dropout ablation for EA and IP prediction on the VIPEA dataset. Values represent mean RMSE ± standard deviation across three random seeds, with predictions pooled across the nine evaluation folds for each seed. All values are reported in eV, with the best-performing variant for each property highlighted in bold.
    }
    \label{tab:motif-vocabulary-dropout}
    \begin{tabular}{lcc}
        \toprule
        \textbf{Motif vocabulary} & \textbf{EA RMSE (eV)} & \textbf{IP RMSE (eV)} \\
        \midrule
        0\% dropout              & $\mathbf{0.070 \pm 0.002}$ & $\mathbf{0.064 \pm 0.003}$ \\
        50\% dropout      & $0.072 \pm 0.003$ & $0.069 \pm 0.004$ \\
        75\% dropout      & $0.077 \pm 0.006$ & $0.079 \pm 0.006$ \\
        \bottomrule
    \end{tabular}
\end{table}

Table~\ref{tab:motif-pooling} compares mean and attention pooling for aggregating atom representations into motif embeddings. Attention pooling yields slightly lower RMSE than mean pooling for both EA and IP, corresponding to relative reductions of 1.2\% and 1.4\%, respectively. The small differences indicate that the choice of pooling method has a limited effect on predictive performance in this evaluation.

\begin{table}[t]
    \centering
    \caption{Comparison of mean and attention pooling strategies for constructing motif embeddings in EA and IP prediction on the VIPEA dataset. Values represent mean RMSE ± standard deviation across three random seeds, with predictions pooled across the nine evaluation folds for each seed. All values are reported in eV.}
    \label{tab:motif-pooling}
    \begin{tabular}{lcc}
        \toprule
        \textbf{Pooling method} & \textbf{EA RMSE (eV)} & \textbf{IP RMSE (eV)} \\
        \midrule
        Mean pooling      & $0.0708 \pm 0.002$ & $0.065 \pm 0.003$ \\
        Attention pooling & $0.070 \pm 0.002$ & $0.064 \pm 0.003$ \\
        \midrule
        Relative reduction & 1.2\% & 1.4\% \\
        \bottomrule
    \end{tabular}
\end{table}

\begin{figure}[h]
    \centering
    \includegraphics[width=\linewidth,
     keepaspectratio]{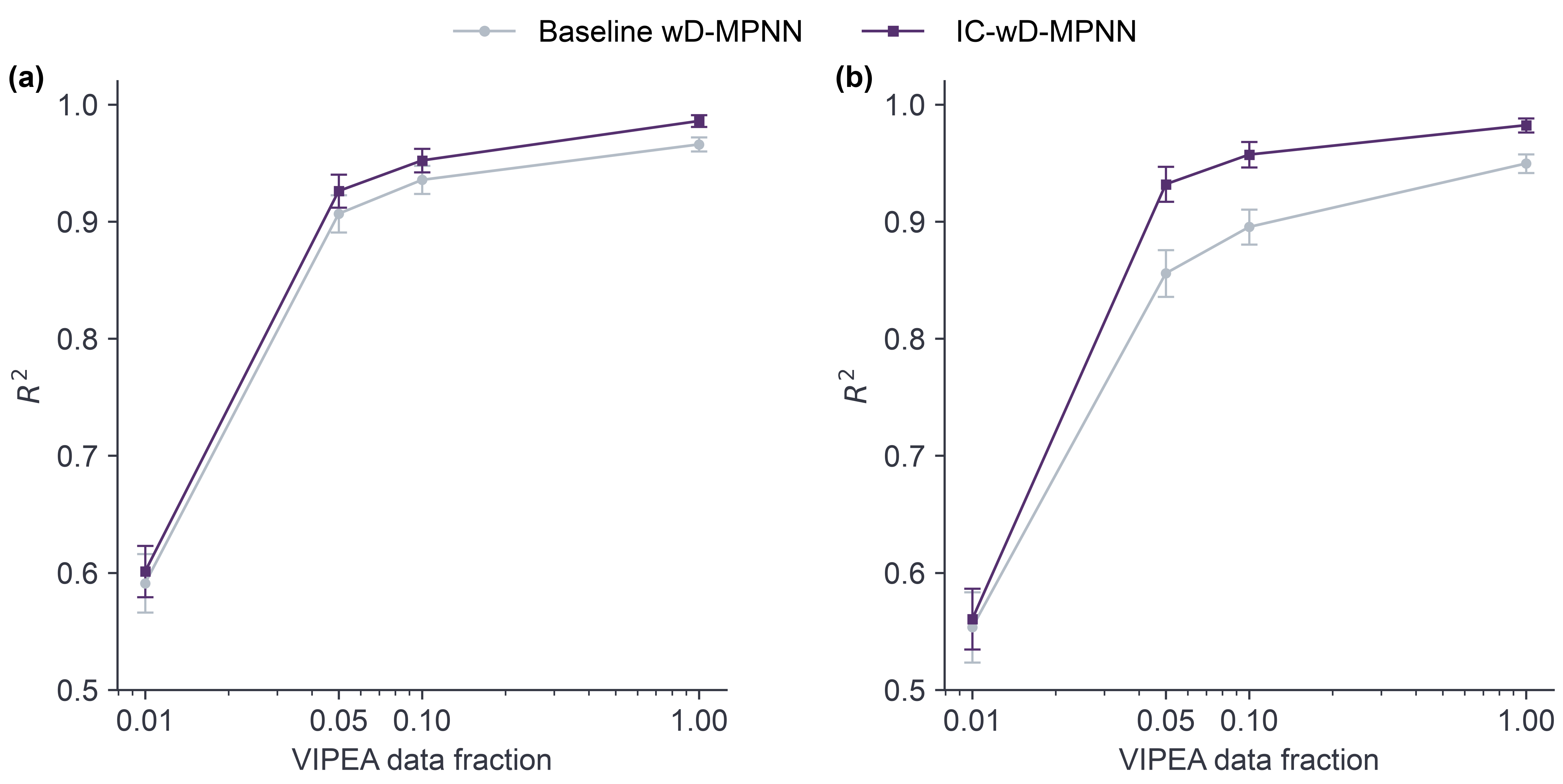}
    \caption
    {Effect of training-set size on EA and IP prediction performance on the VIPEA dataset. $R^2$ is reported for the baseline wD-MPNN and the proposed formulation using 1\%, 5\%, 10\%, and 100\% of the training data.}
    
    \label{fig:training_data_fraction}
\end{figure}

Figure~\ref{fig:training_data_fraction} compares the $R^2$ achieved by the baseline and the proposed formulation across different fractions of the VIPEA training data. The proposed formulation achieves higher $R^2$ than the baseline at all evaluated data fractions for both EA and IP prediction. The improvement is particularly pronounced for IP at intermediate training fractions (5\% and 10\%), whereas the difference is smaller for EA. Both models show substantial improvements as the training set increases, indicating the benefit of additional training data. Overall, these results suggest that the full formulation makes more effective use of limited training data, particularly for IP prediction.

\vspace{5mm}

Table~\ref{tab:model-comparison} sets the baseline wD-MPNN update equations alongside the proposed formulation, row by row. Stage~I is the baseline wD-MPNN update rule applied over layers $0\!\to\!K-1$. Stage~II applies the same rule over layers $K\!\to\!D-1$, except that the skip connection uses the regenerated messages $\widetilde{\mathbf{m}}_{ij}^{(K)}$ in place of the initial messages $\mathbf{m}_{ij}^{0}$; the atom aggregation and polymer readout are unchanged. For clarity, directed-edge messages are denoted by \(\mathbf{h}_{vu}\) in the baseline formulation and by \(\mathbf{m}_{ij}\) in the proposed formulation. Both represent the same type of directed-edge message; the difference in notation is introduced solely to distinguish the two formulations. The four rows marked as absent from the baseline constitute the contribution we present here: attention pooling of contextual atom embeddings into motif representations $\mathbf{z}_m$, addition of $\mathbf{z}_{\phi(i)}$ to the representations of interface atoms, and regeneration of the outgoing messages from those atoms alone. The attention weights are given here in full, having been abbreviated as $\mathrm{AttnPool}$ in Eq.~\ref{eq:motif-pooling} in the main text. The colour coding distinguishes the components inherited from the baseline wD-MPNN formulation from those introduced by IC-wD-MPNN. Slate grey denotes the baseline formulation, highlighting the mathematical components retained from the original architecture, while deep purple identifies the proposed formulation and its modifications, making it explicit where the two formulations share the same mathematical structure and where IC-wD-MPNN introduces additional operations for 
interface conditioning.

\newpage

\begin{table*}[b!]
\centering
\footnotesize
\renewcommand{\arraystretch}{2.0}
\setlength{\tabcolsep}{8pt}

\caption{Comparison between the mathematical formulation of the baseline wD-MPNN  and the proposed IC-wD-MPNN architecture}
\label{tab:model-comparison}
\vspace{1mm}
\makebox[\textwidth][c]{%
\begin{tabular}{
    >{\centering\arraybackslash}m{0.47\textwidth}
    >{\centering\arraybackslash}m{0.47\textwidth}
}

\toprule

\baselineheading{Baseline wD-MPNN}
&
\motifheading{Proposed IC-wD-MPNN}
\\

\midrule

\multicolumn{2}{c}{\baselineheading{Initialization}}
\\

\midrule

$\displaystyle
\begin{aligned}
\mathbf{h}_{vu}^{0}
&=
\tau\left(
\mathbf{W}_i\,
\mathrm{cat}(x_v,e_{vu})
\right)
\end{aligned}
$

&
$\displaystyle
\begin{aligned}
\mathbf{m}_{ij}^{0}
&=
\tau\left(
\mathbf{W}_i\,
\mathrm{cat}(\mathbf{x}_i,\mathbf{e}_{ij})
\right)
\end{aligned}
$

\\[4mm]

\midrule

\multicolumn{2}{c}{\baselineheading{Stage I: Initial message passing}}
\\

\midrule

$\displaystyle
\begin{aligned}
\mathbf{h}_{vu}^{t+1}
&=
\tau\left(
\mathbf{h}_{vu}^{0}
+
\mathbf{W}_h
\sum_{k\in N(v)\setminus u}
w_{kv}\mathbf{h}_{kv}^{t}
\right),\\
&\qquad t=0,\ldots,D-1
\end{aligned}
$

&

$\displaystyle
\begin{aligned}
\mathbf{m}_{ij}^{t+1}
&=
\tau\left(
\mathbf{m}_{ij}^{0}
+
\mathbf{W}_h
\sum_{k\in N(i)\setminus j}
w_{ki}\mathbf{m}_{ki}^{t}
\right),\\
&\qquad t=0,\ldots,K-1
\end{aligned}
$

\\[4mm]

\midrule

\multicolumn{2}{c}{\motifheading{Interface conditioning module}}
\\

\midrule

\parbox[c]{0.85\linewidth}{\centering
\textit{Not present in the baseline wD-MPNN}
}
&
\parbox[c]{0.95\linewidth}{\centering
$\displaystyle
\begin{gathered}
\text{\modulelabel{Intermediate atom representations}}\\[-0.5mm]
\begin{aligned}
\mathbf{h}_i^{(K)}
&=
\tau\left(
\mathbf{W}_o
\mathrm{cat}
\left(
\mathbf{x}_i,
\sum_{k\in\mathcal{N}(i)}
w_{ki}\mathbf{m}_{ki}^{(K)}
\right)
\right)
\end{aligned}
\end{gathered}
$
}

\\[8mm]

\parbox[c]{0.85\linewidth}{\centering
\textit{Not present in the baseline wD-MPNN}
}
&
\parbox[c]{0.95\linewidth}{\centering
$\displaystyle
\begin{gathered}
\text{\modulelabel{Motif embedding construction}}\\[-1mm]
\begin{aligned}
a_i
&=
\mathbf{w}^{\top}
\tanh(\mathbf{W}_a\mathbf{h}_i^{(K)}),\\
\alpha_i
&=
\frac{\exp(a_i)}
{\sum_{j\in M_m}\exp(a_j)},\\
\mathbf{z}_m
&=
\sum_{i\in M_m}
\alpha_i\mathbf{h}_i^{(K)}
\end{aligned}
\end{gathered}
$
}

\\[12mm]

\parbox[c]{0.85\linewidth}{\centering
\textit{Not present in the baseline wD-MPNN}
}
&
\parbox[c]{0.95\linewidth}{\centering
$\displaystyle
\begin{gathered}
\text{\modulelabel{Interface atom enrichment}}\\[0.5mm]
\begin{aligned}
\widetilde{\mathbf{h}}_i^{(K)}
&=
\mathbf{h}_i^{(K)}
+
\mathbf{z}_{\phi(i)}
\end{aligned}
\end{gathered}
$
}

\\[4mm]

\parbox[c]{0.85\linewidth}{\centering
\textit{Not present in the baseline wD-MPNN}
}
&
\parbox[c]{0.95\linewidth}{\centering
$\displaystyle
\begin{gathered}
\text{\modulelabel{Outgoing message regeneration}}\\[1mm]
\begin{aligned}
\widetilde{\mathbf{m}}_{ij}^{(K)}
&=
\begin{cases}
\tau\left(
\mathbf{W}_r
\mathrm{cat}
\left(
\widetilde{\mathbf{h}}_i^{(K)},
\mathbf{e}_{ij}
\right)
\right),
& i\in\mathcal{I},
\\[2mm]
\mathbf{m}_{ij}^{(K)},
& i\notin\mathcal{I}.
\end{cases}
\end{aligned}
\end{gathered}
$
}

\\[4mm]

\midrule

\multicolumn{2}{c}{\baselineheading{Stage II: Post-conditioning message passing}}
\\

\midrule

\textit{Already completed in Stage I}

&

$\displaystyle
\begin{aligned}
\widetilde{\mathbf{m}}_{ij}^{(t+1)}
&=
\tau\left(
\widetilde{\mathbf{m}}_{ij}^{(K)}
+
\mathbf{W}_h
\sum_{k\in\mathcal{N}(i)\setminus j}
w_{ki}\widetilde{\mathbf{m}}_{ki}^{(t)}
\right),
\\[-1mm]
&\qquad t=K,\ldots,D-1
\end{aligned}
$

\\[4mm]

\midrule

\multicolumn{2}{c}{\baselineheading{Final atom aggregation}}
\\

\midrule

$\displaystyle
\begin{aligned}
\mathbf{h}_v
&=
\tau\left(
\mathbf{W}_o
\mathrm{cat}
\left(
\mathbf{x}_v,
\sum_{k\in N(v)}
w_{kv}\mathbf{h}_{kv}^{D}
\right)
\right)
\end{aligned}
$

&

$\displaystyle
\begin{aligned}
\mathbf{h}_i^{(D)}
&=
\tau\left(
\mathbf{W}_o
\mathrm{cat}
\left(
\mathbf{x}_i,
\sum_{k\in\mathcal{N}(i)}
w_{ki}\widetilde{\mathbf{m}}_{ki}^{(D)}
\right)
\right)
\end{aligned}
$

\\[4mm]

\midrule

\multicolumn{2}{c}{\baselineheading{Polymer readout}}
\\

\midrule

$\displaystyle
\begin{aligned}
\mathbf{h}
&=
\frac{1}{|V|}
\sum_{v\in G}w_v\mathbf{h}_v
\end{aligned}
$

&

$\displaystyle
\begin{aligned}
\mathbf{h}
&=
\frac{1}{|V|}
\sum_{i\in G}w_i\mathbf{h}_i^{(D)}
\end{aligned}
$

\\

\bottomrule

\end{tabular}%
}
\end{table*}

\end{document}